\documentclass[11pt]{article}%

\usepackage{url}
\usepackage{multirow}
\usepackage{array}

\usepackage{mathtools}%
\usepackage{xcolor}
\usepackage{amssymb}
\usepackage{amsmath}
\usepackage{latexsym}
\usepackage[a4paper, left=2cm, right=2cm, top=3cm, bottom=3cm]{geometry}%

\usepackage[titletoc,title]{appendix}%
\usepackage[utf8]{inputenc}%
\usepackage{dsfont}%
\usepackage{times}
\usepackage{soul}
\usepackage[hidelinks]{hyperref}
\usepackage[utf8]{inputenc}
\usepackage[small]{caption}
\usepackage{algorithm} 
\usepackage{algpseudocode}
\usepackage{subcaption}
\usepackage{amsthm}
\usepackage{booktabs}
\usepackage{comment}
\usepackage{afterpage}
\usepackage{xtab}
\usepackage{fullpage}

\usepackage[round]{natbib}

\newcommand{\discri}{{\bf D}}
\newcommand{\gene}{{\bf G}}
\newcommand{\rfirst}[1]{\textbf{#1}}
\newcommand{\rsecond}[1]{{\it \underline{#1}}}
\newcommand{\rlast}[1]{{\it {#1}}}
\newcommand{\synth}{SYNTHIA }

\renewcommand*{\thefootnote}{\fnsymbol{footnote}}

\begin{document}

\title{Adversarial Semi-Supervised Domain Adaptation for Semantic Segmentation: A New Role for Labeled Target Samples}
\maketitle
\author{
\begin{tabular}{*{4}{>{\centering}p{.45\textwidth}}}
Marwa Kechaou  \footnote             				& Mokhtar Z. Alaya  \tabularnewline
{\it LITIS UR 4108, F-76000}						& {\it LMAC EA 2222, F-60200} \tabularnewline  
{\it INSA Rouen Normandie}                       	& {\it Université de Technologies de Compiègne} \tabularnewline
Rouen, France                      	                & Compiègne, France \tabularnewline
\texttt{marwa.kechaou@insa-rouen.fr} 		& \texttt{alayaelm@utc.fr} \\ \tabularnewline 

Romain Hérault  							& Gilles Gasso \tabularnewline
{\it GREYC, F-14000}               						& {\it LITIS UR 4108, F-76000}\tabularnewline 
{\it Université de Caen Normandie}               	& {\it INSA Rouen Normandie} \tabularnewline
Caen, France                      	                & Rouen, France \tabularnewline
\texttt{romain.herault@unicaen.fr} 			& \texttt{gilles.gasso@insa-rouen.fr} \tabularnewline
\end{tabular}
}

\begin{abstract}
Adversarial learning baselines for domain adaptation (DA) approaches in the context of semantic segmentation are under explored in semi-supervised framework. 
These baselines involve solely the available labeled target samples in the supervision loss.
In this work, we propose to enhance their usefulness on both semantic segmentation and the single domain classifier neural networks.
We design new training objective losses for cases when labeled target data behave as source samples or as real target samples. 
The underlying rationale is that considering the set of labeled target samples as part of source domain helps reducing the domain discrepancy and, hence, improves the contribution of the adversarial loss.
To support our approach, we consider a complementary method that mixes source and labeled target data, then applies the same adaptation process. 
We further propose an unsupervised selection procedure using entropy to optimize the choice of labeled target samples for adaptation. 
We illustrate our findings through extensive experiments on the benchmarks GTA5, SYNTHIA, and Cityscapes. The empirical evaluation highlights competitive performance of our proposed approach. 
\end{abstract}

\footnotetext{ Corresponding author.}
\let\thefootnote\relax\footnotetext{This paper is currently under consideration at Computer Vision and Image Understanding Journal.}

\section{Introduction}
\noindent Semantic segmentation aims at classifying each pixel of a given image to provide a semantic visual description of the image content. The vast majority of the segmentation models is based on deep networks with  several  applications in various  fields. 
Generally, semantic segmentation task requires a tedious annotation of the training images. To alleviate the problem, research works have leveraged transfer learning \citep{Sun_2019_CVPR} or adaptation \citep{tsai2018learning} principles to adjust a segmentation network learned on a certain domain called source to a new domain coined target. In that regard, unsupervised domain adaptation (UDA) methods for semantic segmentation have been widely explored to narrow the domain gap between labeled source images and unlabeled target ones \citep{tsai2018learning,vu2019advent}. 
However, although these procedures help improve the segmentation performances in target domain, their achieved performances are still far from those of {\it oracle models} {, i.e.} using supervised training on a large set of labeled target images.

\noindent To overcome these limitations, recent works have alleviated the UDA to the Semi-supervised domain adaptation (SSDA) framework  \citep{wang2020alleviating,liu2021domain,chen2021semi}. The goal is to use a small set of  labeled target images and an amount of unlabeled target images to guide the domain matching.
\cite{wang2020alleviating} applied adversarial learning and used only the source and unlabeled target data  to minimize the domain discrepancy for the global adaptation.   In  contrast, \cite{liu2021domain,chen2021semi} proposed alternative domain matching methods to improve the performance of UDA and to prove the effectiveness of their approaches compared to the adversarial procedure where labeled target images were exclusively restricted to the segmentation loss. Hence, the use of the target labeled data is under-explored in the adversarial learning for SSDA.

\noindent Hereafter, we focus on an
adversarial approach to address SSDA for semantic  image segmentation. We propose different methods to handle the labeled target data in the adversarial learning  scheme to enhance the standard approach. 
In classical adversarial DA, the training procedure consists of a segmentation network that outputs semantic maps, and of a discrimination network that learns to correctly classify the domain of input maps, i.e source or target. During the training, the  segmentation model learns to reduce the drift between source and target domains and predict maps with similar distributions to fool the discriminator.

\noindent The main challenge of adversarial models is to maintain the same learning speed for both networks~\citep{goodfellow2014generative}. A common issue is to have an earlier superior discriminator in the competition since its task is much easier. Papers such as WGAN \citep{arjovsky2017wasserstein} have proposed alternatives to ensure learning stability by suggesting a new critic for the discriminator. 

\begin{figure}[H]
\centering
\includegraphics[width=1.\textwidth]{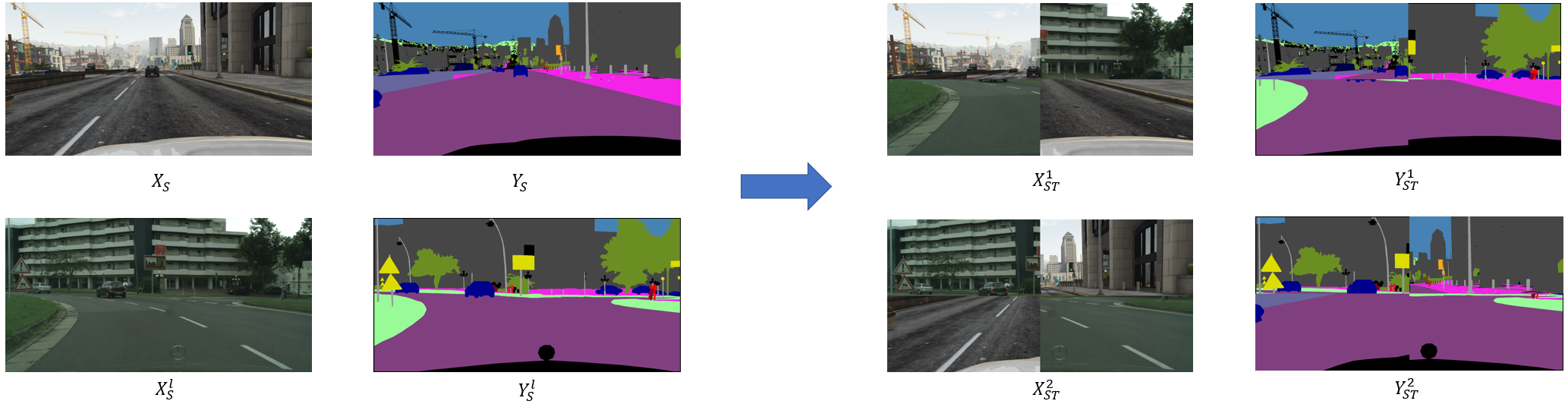}
\caption[]{Example of mixed source and labeled target inputs.}
\label{fig:real_mixed_patches}
\end{figure}

\noindent In this work, we devise a new adversarial learning strategy adapted to  SSDA. Namely, we create an auxiliary source domain that regroups source and labeled target data and thus harden the discriminator task which is required to guess labeled versus unlabeled images rather than source vs target samples. The overall learning principle is illustrated in Figure \ref{fig:architecture}. Besides, we provide the discriminator  mixed  patches of images from source and  target labeled samples as in Fig.~\ref{fig:real_mixed_patches}. This augmentation scheme generates a diversity in the inputs from source and target domains and helps the discriminator in providing relevant feedback to the segmentation network. To show the effectiveness of the approach, we experimentally  highlight its potentials in comparison to the case where labeled target images solely act as target data in the adversarial scheme.

\noindent When addressing SSDA, an issue is the  selection of the target samples to  be annotated for a better adaptation.
Towards this end, we rely on the entropy function applied to the maps, that  determines the most challenging target inputs for the segmentation network trained solely on the source data. We also investigate random selection procedure of the labeled target samples to show some variability in results and demonstrate the importance of entropy selection.

\medskip

\noindent The main contributions are summarized as follows:
\begin{itemize}
    \item We design new training objective losses for the adversarial baseline architecture in SSDA framework without additional networks and hyperparameters. 
    \vspace{-0.1cm}
    \item We exploit entropy uncertainty criteria to propose an unsupervised selection procedure for challenging target domain samples to be annotated. 
    \vspace{-0.1cm}
    \item We conduct extensive numerical experiments showing the efficiency of the new role assigned to labeled target samples as well as the entropy selection procedure.
\end{itemize}

\section{Related Work}
\paragraph{Unsupervised Domain Adaptation} Approaches for UDA seek the best match between drifted data distributions without any target supervision.
For classification tasks, DANN~\citep{ganin2015unsupervised}  and CDAN~\citep{long2017conditional} use a domain classifier to align the distributions of feature domains, either by reverse gradient via back-propagation or by matching multimodal features.
RTN~\citep{long2016unsupervised}, MCD~\citep{saito2018maximum} and ADR~\citep{saito2017adversarial} break with the common assumption of having a single classifier and use multiple classifiers.
The optimization of minimax problem is applied to target data predictions.
RTN minimizes the residual loss differentiating the classifiers to match their marginal distributions contrarily to MCD and ADR that maximize their discrepancy to superpose feature domain supports.

\noindent In semantic segmentation, adversarial approaches, such as AdaSeg~\citep{tsai2018learning} and AdvEnt~\citep{vu2019advent}, apply distribution matching to the output space (segmentation maps).
Indeed, the decomposition of fully convolutional networks into feature extractor and classifier often remains ambiguous. 
UDA approaches mainly involve entropy minimization \citep{vu2019advent} or squared loss maximization to unlabeled samples~\citep{chen2019domain}.
Recently, self-supervised learning has been widely used and there are a variety of methods for generating pseudo-labels of unlabeled data~\citep{zou2018unsupervised,zou2019confidence,huang2022multi}.

\paragraph{Semi-Supervised Domain Adaptation} It assumes the availability of some labeled target data.
In image classification, SSDA approaches apply different strategies of feature alignment using the main network task.
MME~\citep{saito2019semi} aligns the source and target features by optimizing the estimation of class prototypes via minimax entropy.
Similarly, APE~\citep{kim2020attract} applies three steps (attraction, perturbation and exploration) to reduce the discrepancy of intra-domain within the target one.
For semantic segmentation, unlike UDA, there is no standard method for SSDA frameworks.
ASS~\citep{wang2020alleviating} proposes an adversarial approach allowing the use of multiple discriminators to alleviate the semantic adaptation and the global shift.
Method in~\cite{liu2021domain} involves self-training via pseudo labels as well as contrastive learning to ensure the alignment of clusters from same classes by matching patches with similar representations from both domains. 
Recent approaches follow Semi-supervised learning (SSL) techniques where labeled and unlabeled samples are drawn from the same distributions contrarily to their definition in DA.
\cite{chen2021semi} apply Multi-teacher Student approach of SSL with region level data mixing as in~\cite{yun2019cutmix}.
All of these SSDA approaches have used different alternatives to conventional adversarial UDA techniques.
In contrast to these previous works, we differently explore the baseline adversarial approach by drastically drifting on how labeled target samples are considered by the discriminator, and show its effectiveness.

\section{Proposed Method}

\paragraph{Notation}
Let $\mathcal{D}_S=\{(x_s^i, y_s^i)\}_{i=1}^{N_s}$ be the source domain labeled samples with $x_s \in \mathbb { R } ^ { H \times W \times d }$ and $y_s \in \{1,..,C\} ^ { H \times W }$  
where $H$ and $W$ correspond to length and width of an input image, respectively, $d$ is the number of its channels, and $C$ stands for the number of classes. 
We note that source and target classes are the same.
We assume that we have partial access to the labels of target domain $\mathcal{D}_T$. We denote  $\mathcal{D}_T^l$ and  $\mathcal{D}_T^{u}$ to be, the sets of labeled and unlabeled target domain samples, respectively. Namely, $\mathcal{D}_T = \mathcal{D}_T^l \cup \mathcal{D}_T^{u}$, where $\mathcal{D}_T^l=\{({x_t^{l,i}}, {y_t^{l,i}})\}_{i=1}^{N_t^l}$,  $\mathcal{D}_T^{u}=\{{x_t^{u,i}}\}_{i=1}^{N_t^u}$, and $N_t = N_t^l + N_t^u.$ 
We denote by $\theta_{\bf G}, \theta_{\bf D}$ the weight parameters of the networks $\bf G, \bf D$, respectively. 
\subsection{Background of Adversarial UDA}
In adversarial UDA ($N_t^l=0$), the DA algorithm uses two networks: semantic segmentation ${\bf G}$ and discriminator ${\bf D}$~\cite{tsai2018learning,vu2019advent}. Adaptation is applied to the output space instead of the feature space, contrarily to the classification~\cite{ganin2015unsupervised}. Let $C$ be the number of classes and 
$P_s \in [0,1] ^ { H \times W \times C}$ and $P_t \in [0,1] ^ { H \times W \times C}$ be the predicted masks by ${\bf G}$ on source and target images respectively.

\noindent In \cite{tsai2018learning}, $P_s$ and $P_t$ are fed into the discriminator ${\bf D}$ which determines the domain of each mask. In contrast, \cite{vu2019advent} rather computes the entropy maps of the masks defined as $\texttt{Entropy}(P) = -P\odot\log(P)$  to propose an adversarial entropy minimization method termed as AdvEnt. Here the operator $\odot$ stands for the elementwise multiplication. In the sequel, we denote by $\mathcal{I} = \texttt{Entropy} \circ {\bf G}$ with $\circ$ representing standard function decomposition. During the training, ${\bf G}$ learns to predict maps $P_t$ whose distribution is close to the distribution of source semantic maps via the adversarial loss. The learning procedure alternates the training of ${\bf G}$ and ${\bf D}$ by applying the following training objectives:
\begin{equation}
\label{lo_G_UDA}
 \min_{\substack{\theta_{\bf G}}} \frac{1}{N_s} \sum_{x_s} {L_{seg}(\gene(x_s),y_s)} + {\frac{\lambda_{adv}}{N_t} {\sum_{x_t}} {L_{dis}(\discri(z_t),S)}}  
\end{equation}
\begin{equation}
\label{lo_D_UDA}
\min_{\substack{\theta_{\bf D}}} \frac{1}{N_s} \sum_{x_s} {L_{dis}(\discri(z_s),S)} + \frac{1}{N_t} \sum_{x_t} {L_{dis}(\discri(z_t),T)}  
\end{equation}

where $z$ is defined as the direct output defined as:
\begin{equation}
\label{def_of_z}
z = \left\lbrace
\begin{array}{lr}
    {\bf G}(x) & \text{\citep{tsai2018learning} }   \\
    \mathcal{I}({x})= \texttt{Entropy} \circ {\bf G} (x) & \text{  \citep{vu2019advent} }
\end{array}
\right.
\end{equation}
$L_{seg}$ denotes the cross entropy loss  and $L_{dis}$ represents the binary cross entropy loss. The parameter $\lambda_{adv}>0$ is the tuning hyperparameter for the adversarial 
loss. Adversarial term in Equation~\eqref{lo_G_UDA} is in bold.
The source label from the discriminator's point of view is denoted by $S$ and the target label by $T$.

\begin{figure*}[htpb]
\centering
\includegraphics[width=1.0\textwidth]{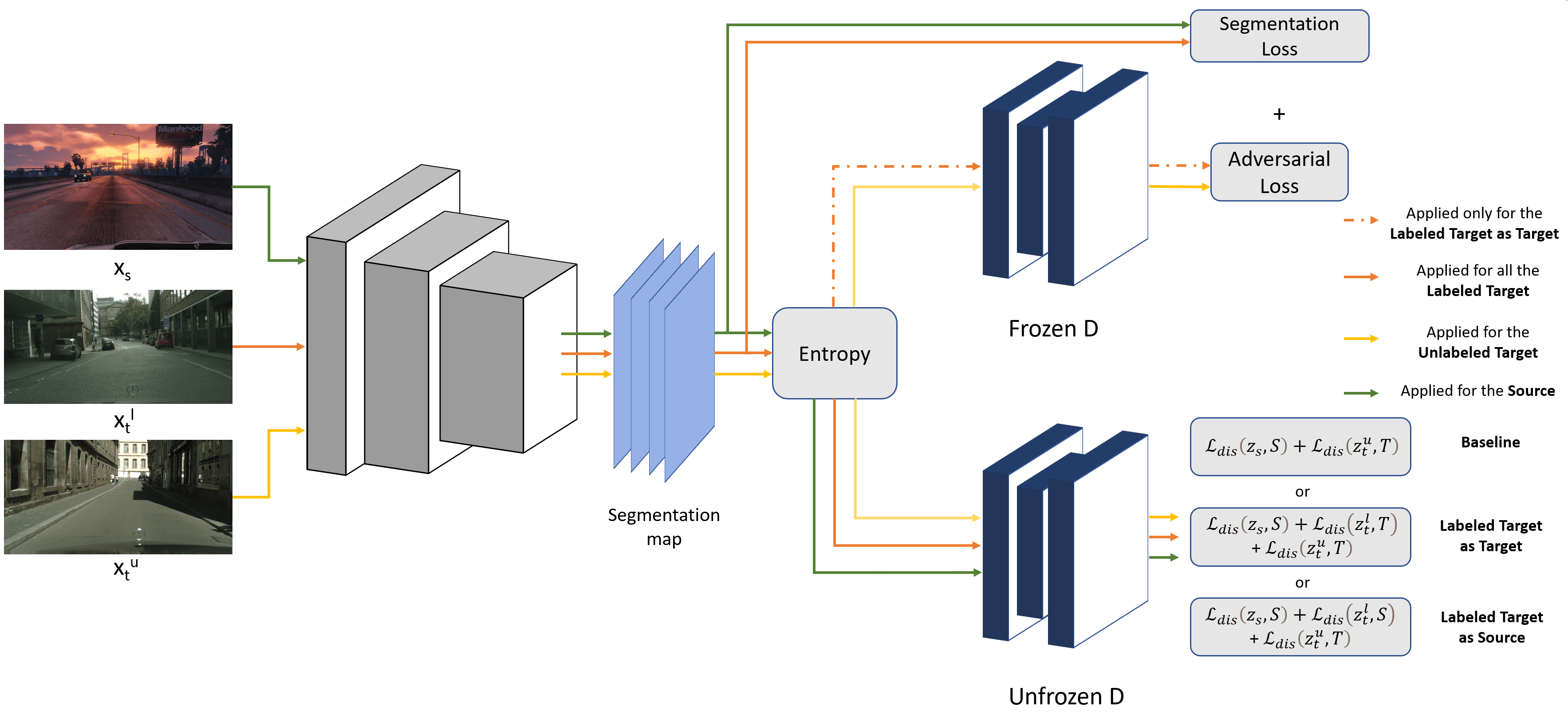}
\caption[]{Illustration of alternative training of segmentation ${\bf G}$ and discriminator ${\bf D}$ networks within an SSDA framework. Source ($x_s$), labeled target ($x_t^l$), and unlabeled target ($x_t^u$) samples are represented by different colors to reveal their contribution in each loss term. The inputs of discriminator {\bf D} are given by Eq.~\eqref{def_of_z}. The loss function of {\bf D} changes according to the role assigned to ($x_t^l$): Labeled Target as Target (see Eq.~\eqref{lo_D_SSDA_2}) or Labeled Target as Source (see Eq.~\eqref{lo_D_SSDA_1}).}
\label{fig:architecture}
\vspace{-0.5cm}
\end{figure*}

\subsection{Adversarial SSDA Framework}
In SSDA, 
~\citep{liu2021domain,chen2021semi} show the superiority of their methods over the semi-supervised form of AdvEnt~\citep{vu2019advent}, where labeled target samples $\{({x_t^l}^i,{y_t^l}^i)\}$ are only used in $L_{seg}$.
Namely, Equations~\eqref{lo_G_UDA} and~\eqref{lo_D_UDA} become  
\begin{multline}
\label{lo_G_SSDA_0} 
\min_{\substack{\theta_{\bf G}}} \frac{1}{N_s} \sum_{x_s} {L_{seg}(\gene(x_s),y_s)} +
\frac{1}{N_t^l} \sum_{x_t^l} {L_{seg}(\gene(x_t^l),y_t^l)}
+{\frac{\lambda_{adv}}{N_t^u} {\sum_{x_t^{u}}} {L_{dis} (\discri({z_t^{u}}),S)}}
\end{multline}
\begin{equation}
\label{lo_D_SSDA_0}
\min_{\substack{\theta_{\bf D}}} \frac{1}{N_s} \sum_{x_s} {L_{dis} (\discri({z_s}),S)} + \frac{1}{N_t^u} {\sum_{x_t^{u}}} {L_{dis} (\discri(z_t^{u}),T)}  
\end{equation}
\noindent  with $z = \mathcal{I}({x})=\texttt{Entropy} \circ {\bf G} (x)$ as in \cite{vu2019advent}.
The labeled samples $x_t^l$ and unlabeled ones $x_t^{u}$ are assigned specific roles: the unlabeled set $\mathcal{D}_T^{u}$ serves for domain gap reduction as classically \cite{ganin2015unsupervised} while the small target labeled set $\mathcal{D}_T^{l}$ is used to guess the segmentation error in the target domain.
We consider Eqs.~\eqref{lo_G_SSDA_0} and~\eqref{lo_D_SSDA_0} as the \textbf{baseline} approach.

\paragraph{Propositions} Starting from that, we devise a new adversarial scheme that may provide an enhanced domain matching feedback to the segmentation network. Indeed, in addition to the segmentation loss, we include the target samples $\{{x_t^l}^i\}$ in training the adversarial loss and the  discriminator ${\bf D}$ under two settings: 

\begin{itemize}
\item[(1)] $\{{x_t^l}^i\}$ belongs to the source domain $\mathcal{D}_S$. Hence, the training objective of ${\bf G}$ remains the same as in Eq.~\eqref{lo_G_SSDA_0}, while that of ${\bf D}$ adds a loss term to classify $\{{x_t^l}^i\}$ as source data, i.e., 
\begin{multline}
\label{lo_D_SSDA_1}
\min_{\substack{\theta_{\bf D}}} \frac{1}{N_s} \sum_{x_s} {L_{dis} (\discri({z_s}),S)} 
+ \frac{1}{N_t^l} {\sum_{x_t^l}} {L_{dis} (\discri({z_t^l}),S)} 
+ \frac{1}{N_t^u} {\sum_{x_t^{u}}} {L_{dis} (\discri({z_t^{u}}),T)}
\end{multline}

\item[(2)] $\{{x_t^l}^i\}$ belong to target domain $\mathcal{D}_T$ and is used in adversarial and discriminator losses. The new training objectives are presented by the following equations: 
\begin{multline}
\label{lo_G_SSDA_2}
\min_{\substack{\theta_{\bf G}}} \frac{1}{N_s} \sum_{x_s} {L_{seg}(\gene(x_s),y_s)}
+\frac{1}{N_t^l} \sum_{x_t^l} {L_{seg}(\gene(x_t^l),y_t^l)} \\
\quad +{\frac{\lambda_{adv}}{N_t^l} {\sum_{x_t^l}} {L_{dis} (\discri({z_t^l}),S)}}
+ {\frac{\lambda_{adv}}{N_t^{u}} {\sum_{x_t^{u}}} {L_{dis} (\discri({z_t^u}),S)}} 
\end{multline}
\begin{multline}
\label{lo_D_SSDA_2}
\min_{\substack{\theta_{\bf D}}} \frac{1}{N_s} \sum_{x_s} {L_{dis}(\discri({z_s}),S)} 
+ \frac{1}{N_t^l} {\sum_{x_t^l}} {L_{dis} (\discri({z_t^l}),T)} 
+ \frac{1}{N_t^{u}} {\sum_{x_t^{u}}} {L_{dis} (\discri({z_t^{u}}),T)}  
\end{multline}

\end{itemize}

\noindent We summarize the different training objectives in Fig.~\ref{fig:architecture}.

\noindent The first setting amounts to form an auxiliary source domain $\mathcal{D}_S^{\text{aux}} = \mathcal{D}_S \cup \mathcal{D}_T^l$ composed of all source and target labeled samples. Therefore, the goal of the segmentation model $\mathbf{G}$  is to enforce the matching of the distributions of the segmentation masks issued from the unlabeled target set $\mathcal{D}_T^u$ and the labeled auxiliary set $\mathcal{D}_S^{\text{aux}}$ by attempting to mislead the discriminator $\mathbf{D}$. Doing so, we expect that the feature learning and pixelwise classification stages of $\mathbf{G}$ may be guided towards learning a mixture of the source and target domain distributions.

\noindent In the second setting, ${\bf D}$ learns to discriminate between  predicted masks produced by ${\bf G}$ on images from the shifted distributions $\mathcal{D}_S$ and $\mathcal{D}_T^u \cup \mathcal{D}_T^l $. Thus, ${\bf D}$ sees more target examples than in the first setting and the Baseline. Therefore, ${\bf D}$ is likely to learn quickly which may constrain the optimization of ${\bf G}$. Intuitively, to help ${\bf D}$ provide relevant feedback to ${\bf G}$ and prevent it from learning too quickly, we consider the labeled target samples as source data.

\subsection{Adaptation via Data Mixing}
\label{mix_section}
To strengthen the learning capacities of the segmentation model ${\bf G}$, we revisit  the case study based on Eqs.~\eqref{lo_G_SSDA_0} and \eqref{lo_D_SSDA_1} by feeding ${\bf G}$ with more challenging images. For the sake, we define another auxiliary source domain denoted as $\mathcal{D}_S^{\text{aux, mix}}$  composed of a mixed combination of $\mathcal{D}_S$ and $\mathcal{D}_T^l$. Specifically $\mathcal{D}_S^{\text{aux, mix}}$ includes crafted images based on the patch fusion of $\{x_s^i\}$ and $\{{x_t^l}^i\}$. Cut and mix techniques are used in semi-supervised learning. They are based either on cutting out small rectangular regions of an image $x_s$ and mix it at the same position in an image $x_t^l$ \citep{devries2017improved}, or on using flexible binary masks to merge images $x_s$ and $x_t^l$ \citep{french2019semi}. \cite{ren2022simple} utilized the latter technique in self-supervised learning to enhance the reliability of pseudo-labels.

\begin{figure}[h]
\centering
\includegraphics[width=0.9\textwidth]{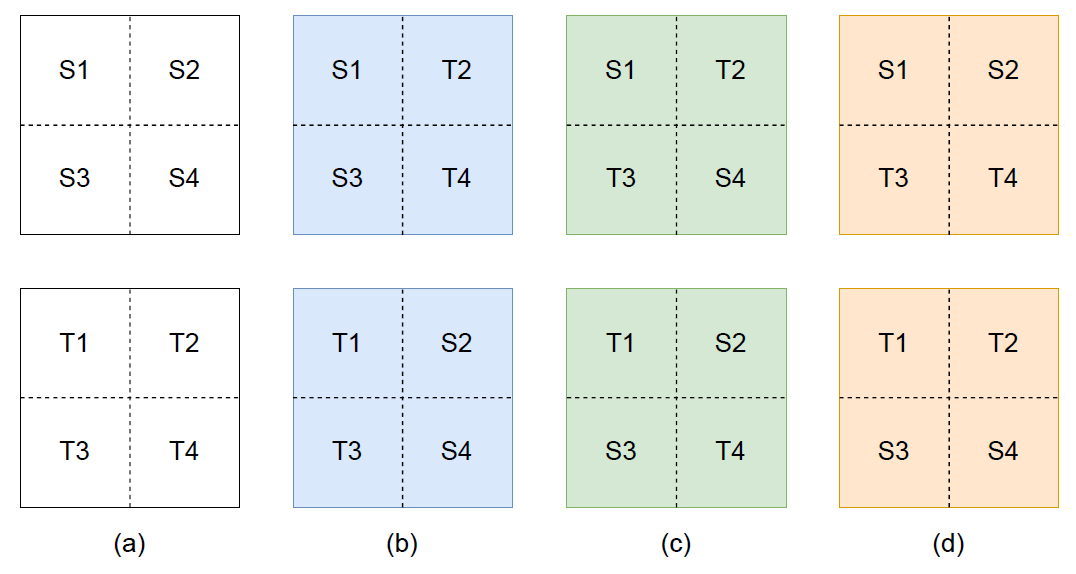}
\caption[]{Different configurations (b), (c), and (d) of source and labeled target input data (a).}
\label{fig:mixed_patches}
\vspace{-0.35cm}
\end{figure}

For our purpose, we first resize the source image to the target image size and then split each input image into 4 patches with equal size as shown in Fig.~\ref{fig:mixed_patches} (a). 
\noindent We mix the patches from source and target images in different configurations (b), (c) and (d). For each configuration, we get two samples representing the new source domain. As a consequence, the predictions of the segmentation network result in the mixing of the masks $P_s$ and $P_t^l$ according to the used configuration. Domain classifier ${\bf D}$ learns to discriminate the mixed predictions as (auxiliary) source inputs. The configurations (a), (b), (c) and (d) are alternatively  applied during the training process, as explained in Algorithm~\ref{algo_mixing} in the Appendix.

\subsection{Choice of Labeled Target Data}
\label{data_choice}
In SSDA framework, it is important to identify the most relevant data to annotate to boost the adaptation.
\cite{saito2019semi} use entropy minimization to penalize samples close to the boundary decision task. Thus, high entropy refers to unreliable predictions. Since we have access to labels of $\mathcal{D}_S$, we divide source data into training and validation sets and then train~${\bf G}$. We calculate the entropy of each target sample $x_t$ based on the predicted map  
$\hat {P}_t =(\hat{p}_t^{j,k,c}) \in [0,1] ^ { H \times W \times C}$:
\begin{align}
\text{Ent}(x_t) =- \sum_{j=1}^{H} \sum_{k=1}^{W} \sum_{c=1}^{C}  \hat{p}^{j,k,c}_t \log(\hat{p}^{j,k,c}_t). \label{equ_003}
\end{align}
We then choose $N_t^l$ target samples with the highest entropy values to be manually labeled. Therefore, we get access to the labels of target samples presenting most ambiguous decisions for ${\bf G}$ pretrained on source data. The remaining target samples form the unlabeled set $\mathcal{D}_T^u$.

\vspace{-0.2cm}
\section{Experiments and Results}

\subsection{Implementation Details}
\paragraph{\textit{Datasets}} 
We conduct experimental evaluation of our domain adaptation approach on GTA5~\cite{richter2016playing}, SYNTHIA-RAND-CITYSCAPES~\cite{ros2016synthia} abbreviated below as \synth\!\!, and Cityscapes~\cite{cordts2016cityscapes} datasets. We consider the following adaptation problems:
\begin{itemize}
\item $\textrm{GTA5}\rightarrow\textrm{Cityscapes}$: GTA5 dataset (source domain) contains $24,966$ images with resolution of $1914\times1052$. Cityscapes (target domain) is composed of $2,975$ images for training and $500$ images for validation. During training, Cityscapes images are resized to $1024\times512$ and GTA5 to $1280\times720$, except for the mixing approach where they are resized like Cityscapes samples. We consider $19$ classes in common for adaptation. 
 \item  \synth  $\rightarrow$ Cityscapes: \synth dataset contains $9,400$ images with resolution of $1280\times760$.  The adaptation is applied to $16$ classes shared between domains.
\end{itemize}

We choose the number of labeled target samples $N_t^l$ in the set $\{100, 200, 500, 1000\}$. We apply the entropy ranking method to annotate target samples as explained in  Section~\ref{data_choice}.

\paragraph{Evaluation metric} To evaluate the adaptation performances we use Mean Intersection over Union (mIoU) reported on the validation set.  For \synth  $\rightarrow$ Cityscapes, the mIoU is  considering $13$ classes as~\cite{tsai2018learning,vu2019advent}.

\paragraph{\textit{Network architectures}} We use Deeplab-V2~\citep{chen2017deeplab} for the segmentation network ${\bf G}$ with the ResNet-101 as a backbone. Adaptation is applied to the multiple outputs of ${\bf G}$ at \textit{conv4} and \textit{conv5} layers. The discriminator ${\bf D}$ has the same structure used in \cite{radford2015unsupervised}. It is a fully convolutional network composed of $5$ convolution layers with kernel size $= 4$,  stride $= 2$ and channels in $ \{64, 128, 256, 512, 1\}$. A Leaky-ReLU  nonlinear activation function follows each of the first four convolution layers.

\paragraph{\textit{Hyperparameters}} We optimize the semantic network $ {\bf G}$, initially pretrained on ImageNet, by minimizing either the training objective \eqref{lo_G_SSDA_0} or \eqref{lo_G_SSDA_2}. As for UDA approaches \citep{tsai2018learning,vu2019advent,chen2019domain}, we use SGD as an optimizer for {\bf G} with initial learning rate $=2.5\times 10^{-4}$, momentum $=0.9$ and weight decay $=5\times 10^{-4}$. 
The total number of iterations is $120, 000$ and we save checkpoints every $2, 000$ iterations as in~\cite{vu2019advent}. 
The discriminator ${\bf D}$ is optimized by minimizing one of the training objectives \eqref{lo_D_SSDA_0}, \eqref{lo_D_SSDA_1} or \eqref{lo_D_SSDA_2}. We use Adam optimizer with learning rate $=10^{-4}$ and momentum $=0.9$ and $0.99.$ The learning rates of {\bf G} and {\bf D} are scheduled according to the polynomial policy $(1 - \frac{iter}{max\_iter})^{0.9}$. As {\bf G} is a Deeplab-V2 having multilevel outputs, we assign $1$ to the main output (from \textit{conv5}) and $0.1$ to the auxiliary output (from \textit{conv4}) in the segmentation loss. Those values are the same whether the input is from $\mathcal{D}_S$ or $\mathcal{D}_T^l$. Accordingly, for adversarial loss, we set $\lambda_{adv}^{main}=10^{-3}$ and $\lambda_{adv}^{aux}=2\times10^{-4}$. The choice of those hyper-parameter values was experimentally argued in~\cite{tsai2018learning}. During the training, we sample $4$ images at each iteration: one from $\mathcal{D}_S$ and  $\mathcal{D}_T^l$ and $2$ from $\mathcal{D}_T^{u}$. The hyperparameters are the same for all the experiments.

\paragraph{\textit{Training process}} We use a single GPU NVIDIA Tesla V100 with 32Go of memory and Pytorch as a deep learning software framework. At each iteration, we update the parameters of the network {\bf G} by minimising segmentation and adversarial losses, while freezing the discriminator weights. Then, the weights of {\bf G} are frozen and the entropy maps of predicted masks on the labeled and unlabeled samples are fed into {\bf D} which is updated to discriminate the domains of its inputs.  
 
\subsection{Results and Analysis}

\noindent We evaluate 5 adaptation methods defined as follows:
\begin{itemize}
\item[(1)] \textbf{Supervised Target (ST)}: is a standard supervised training only on $N_t^l$ samples.
\item[(2)] \textbf{Baseline}: applies Eqs.~\eqref{lo_G_SSDA_0} and ~\eqref{lo_D_SSDA_0} such that the use of $N_t^l$ samples is limited to segmentation loss. 
\item[(3)] \textbf{Labeled Target as Target (LTT)}: includes the labeled samples in the adversarial and  the discriminator losses following Eqs. \eqref{lo_G_SSDA_2} and \eqref{lo_D_SSDA_2}. 
\item[(4)] \textbf{Labeled Target as Source (LTS)}: applies Eqs.~\eqref{lo_G_SSDA_0} and~\eqref{lo_D_SSDA_1}. Here, $N_t^l$ samples behave as source samples. 
\item[(5)] \textbf{Labeled Target as Source - Mix (LTS-Mix)}: is an extension of {\bf LTS} using variable mixed samples as explained in Section~\ref{mix_section}. 
\end{itemize}

\begin{table*}[htpb] 
\setlength{\tabcolsep}{4pt}
\setlength\aboverulesep{0pt}\setlength\belowrulesep{0pt}
\caption{Results on GTA5 $ \rightarrow$ Cityscapes and \synth $\rightarrow$ Cityscapes for different $N_t^l$ values. The first block  presents obtained performances by adversarial UDA approaches using only ${\bf G}$ and ${\bf D}$ as networks ($N_t^l = 0$). Second block reports results from two competitors papers on SSDA. ST refers to supervised semantic segmentation applied to $N_t^l$ target samples i.e. no use of unlabeled samples. Last, performances achieved by our proposed methods with different settings for target sample labeling are presented. Best mIoU per column is in bold font and the second best in italic and underlined.}
\vspace{-0.5cm}
\begin{center}
\begin{tabular}{|c|ccccc||ccccc|}
\cline{2-11}
\multicolumn{1}{c|}{} & \multicolumn{5}{c||}{GTA5 $ \rightarrow$ Cityscapes} & \multicolumn{5}{c|}{\synth $\rightarrow$ Cityscapes}\\
\hline
$N_t^l$ & 0 & 100 &  200 & 500 & 1000 & 0 & 100 &  200 & 500 & 1000\\
\hline
AdaSeg \citep{tsai2018learning} & 42.4 & - & - & - & - & 46.7 & - & - & - & - \\
AdvEnt \citep{vu2019advent}  & 43.8 & - & - & - & -  & 48 & - & - & - & - \\
\hline 
ASS \citep{wang2020alleviating} & - & 54.20 & 56.00 & 60.20 & 64.5 & - & \rsecond{62.10} & 64.80 & 69.8 & 73.00 \\
\citep{liu2021domain} & - & \rfirst{55.17} & \rsecond{56.96} & \rsecond{60.43} & \rfirst{64.62} & - & \rfirst{63.39} & \rsecond{65.23} & \rsecond{70.26} & \rfirst{73.09} \\
\hline
ST & - & 44.24 & 48.72 & 56.43 & 59.75 & - & 54.59 & 61.04 & 65.93 & 68.57\\
\hline 
Baseline & - & 53.22 & 53.41 & 57.63 & 60.52 & - & 58.24 & 60.64 & 64.18 & 67.49\\
LTT & - & 49.72 & 50.83 & 57.01 &  59.95 & - & 57.73 & 60.07 & 63.95 & 67.15 \\
LTS & - & 53.36 & 54.58 & 57.82 & 60.83 & - & 58.41 & {61.78} & {66.23} & {69.12}\\
LTS-Mix & - & \rsecond{54.8} & \rfirst{57.35} & \rfirst{61.28} & \rsecond{64.6} & - & 60.39 & \rfirst{65.48} & \rfirst{70.51} & \rsecond{73.05} \\
\bottomrule
\end{tabular}
\end{center}
\label{table:results_gta_city_synt_city} 
\setlength{\tabcolsep}{1.pt}
\end{table*}

\paragraph{Assessing the Outcomes of Proposed SSDA Algorithms}
\noindent 
Table~\ref{table:results_gta_city_synt_city} reports results of adaptation from GTA5 $\rightarrow$ Cityscapes and SYNTHIA $\rightarrow$ Cityscapes.
For the first adaptation scenario (GTA5 $\rightarrow$ Cityscapes), we notice that almost all the adversarial study cases outperform by a great margin state-of-the-art adversarial UDA models as well as {\bf ST} even as $N_t^l$ increases. It is worth noting that our methods achieve better performances when the target labeled samples $\{x_t^l\}$ behave as source data.
To better understand the effectiveness of the  role assigned to $N_t^l$ samples, we conduct an ablation study by comparing {\bf LTT} and {\bf LTS} to {\bf Baseline}. Indeed, in each setting, $\{x_s\}$ and $\{x_t^u\}$ maintain the same role while the use of $\{x_t^l\}$ changes. In the {\bf Baseline}, $\{x_t^l\}$ contribute exclusively to segmentation loss. In {\bf LTT} and {\bf LTS} settings, $\{x_t^l\}$ contribute both to adversarial loss and discriminator learning. In most cases, the {\bf Baseline} exceeds {\bf LTT}, in particular the difference amounts to more than $2.5\%$ in some experiments (for $N_t^l = 100$ and $200$).  
Intuitively, the discriminator sees more samples from target domain than in {\bf Baseline}, which speeds up its learning and impacts negatively its feedback for the semantic segmentation network. However, our proposed approach {\bf LTS} achieves better performance than {\bf Baseline}
, that confirms its usefulness. We note that {\bf LTS-Mix}, extension of {\bf LTS} by mixing patches of images from $\mathcal{D}_S$ and $\mathcal{D}_T^l$, improves further the results. This confirms the success of the role assigned to the labeled target samples to join the source domain. Indeed, the variety of inputs (4 configurations as shown in Fig.~\ref{fig:mixed_patches}) allows semantic segmentation network and discriminator to learn on mixed domain information and yields an easier distribution alignment. 

\noindent Results of \synth $\rightarrow$ Cityscapes adaptation are shown on the right-hand side of Table~\ref{table:results_gta_city_synt_city}.
The \synth source domain has more diverse views than GTA5, which has an impact on our SSDA algorithm's behavior and its results. In fact, {\bf Baseline} and {\bf LTT} beat only the {\bf ST} for $N_t^l=100$. 
However, {\bf LTS} and {\bf LTS-Mix} keep the best results compared to other study cases. 

\paragraph{Relation to Other SSDA Approaches}
\noindent We report results from \cite{wang2020alleviating} and \cite{liu2021domain} papers in Table~\ref{table:results_gta_city_synt_city}.
We obtain better results than \cite{wang2020alleviating}, which applies adversarial learning through multiple discriminators, for the two adaptation scenarios except for $N_t^l = 100$. 
We also improve results obtained by \cite{liu2021domain} which involves self-training and contrastive learning methods for both adaptation tasks except for $N_t^l = 100$ and in particular for $N_t^l = 1000$, the difference of mIoU performance is less than $0.05\%$. 

\noindent As a conclusion, we empirically show that we can improve adaptation performances by combining $\mathcal{D}_S$ and $\mathcal{D}_T^l$ in one source domain $\mathcal{{D}}_S^{\text{aux}}$ . This could be explained by the fact that domain shift between $\mathcal{{D}}_S^{\text{aux}}$ and $\mathcal{D}_T$ is somehow inferior than the one between $\mathcal{D}_S$ and $\mathcal{D}_T$. 
Additionally, the discriminator {\bf D} struggles to more distinguish source from target predicted mask, by preventing its quick learnability to give relevant information for adjusting the gap between source and target domains.

\begin{figure}[!t]
    \centering
    \begin{subfigure}{0.55\textwidth}
        \centering
        \includegraphics[width=1\linewidth]{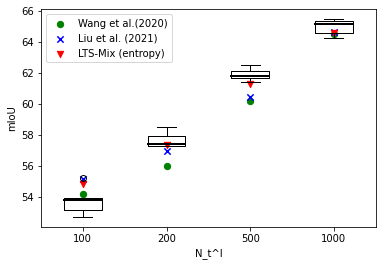}
        \caption[]{GTA $\rightarrow$ Cityscapes}
        \label{fig:box_plot_gta_cityscapes}
    \end{subfigure}

    \begin{subfigure}{0.55\textwidth}
        \centering
        \includegraphics[width=1\linewidth]{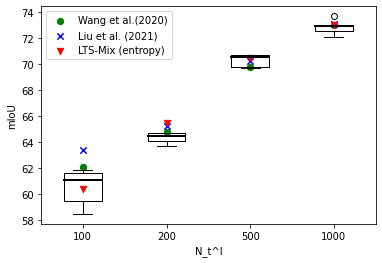}
        \caption[]{\synth $\rightarrow$ Cityscapes}
        \label{fig:box_plot_synthia_cityscapes}
    \end{subfigure}
    \caption{Box plots illustrating variability of {\bf LTS-Mix} mIoU for randomly selected $N_t^l$ samples under $5$ different seeds. The line inside each box represents the median mIoU. The whiskers and points beyond indicate the range of mIoU values. 
    {\bf LTS-Mix} with entropy, \cite{wang2020alleviating}, and \cite{liu2021domain} performance results are added for comparison.
    }
    \label{fig:box_plot}
\end{figure}

\paragraph{Seed Sensitivity Analysis for Choosing Labeled Target Samples}
We evaluate our proposed case studies using random selection strategy of $N_t^l$ samples instead of entropy ranking selection. We then analyze the effect of a chosen seed to evidence that two seeds can have a sensitive difference as given in Fig.~\ref{fig:box_plot}. 
Indeed, we apply the adaptation approach \textbf{LTS-Mix} and vary the \texttt{seed} $\in \{10, 20, 30, 40,50\}$ to randomly select labeled target data $N_t^l \in \{100, 200, 500, 1000\}$.  
We report detailed results for all the cases studied in Tables~\ref{table:results_gta_city_seeds} and~\ref{table:results_synthia_city_seeds}. 
Fig.~\ref{fig:box_plot} shows that the obtained mIoU performances  are strictly dependent on the choice of the random seed. Such a phenomenon was pointed out by~\cite{picard2021torch} who studied in deeply the variations and instabilities related to random seeds selection leading to differences that are significant for the computer vision community. 
We claim that the performances variability for low $N_t^l$ is large. 
Indeed, the difference between best and last best mIoU for $N_t^l=100$ amounts to $2.5\%$ and $3.4\%$ for GTA5 $\rightarrow$ Cityscapes and \synth $\rightarrow$ Cityscapes, respectively. 
For $N_t^l \in \{200,500,1000\}$, these variabilities narrow down.
Due to time and computation budget constraints, we were not able to test for more \texttt{seed} and $N_t^l$ values. Nevertheless, one may notice that performances when considering the mean and std values of \textbf{LTS-Mix} (see Tables~\ref{table:results_gta_city_seeds} and~\ref{table:results_synthia_city_seeds}) on both adaptation scenarios converge to those obtained by their counterparts using entropy-based target samples labeling. 
We emphasize that the best results obtained for {\bf LTS-Mix} with random selection where $N_t^l \in \{500,1000\}$, outperform related work provided for unknown seeds (see Fig.~\ref{fig:box_plot}). This highlights the effectiveness of our approach for both selection strategies.
Overall, it might be unfeasible to find the appropriate \texttt{seed}-value for each $N_t^l$.
As a by-product of this experiment, selecting labeled target data with entropy ranking appears to be the preferred approach, since it identifies the most difficult target images being adapted to source data.
To the best of our knowledge, we are the first to provide precise details on sample selection for SSDA approaches and analyze their seed sensitivity.
\section{Conclusion}
\noindent In this work, a new role for labeled target data is explored for SSDA adversarial approaches. We further confirm this intuitive role through mixing method using simple configurations and reduce the results variability by proposing an entropy ranking selection method. 
Empirical evaluations are conducted on standard DA benchmarks for semantic segmentation. They show that our approach achieved competitive results over other methods requiring more networks or adaptation procedures. 
Our work can lead to important future lines: (i) investigating the entropy selection in Active Learning for Domain Adaptation.  (ii) providing theoretical insights supporting the empirical behavior of LTS-Mix approach by comparing the distributions of the mixed source and target domains using optimal transport metrics. 

\section*{Acknowledgments}
\noindent Marwa Kechaou's work was supported by the National Research Fund of Luxembourg "Fonds National de la Recherche" (FNR).
\noindent Mokhtar Z. Alaya's work has been supported by the Research Direction of University of Technology of Compiègne under the Junior Researchers Program "UTC – Appel soutien Jeunes MC". 
We thank CRIANN (Normandy, France) for the GPU computation facilities.

\bibliographystyle{model2-names}
\bibliography{refs}

\clearpage
\appendix

\section{Data Mixing Algorithm for LTS-Mix Evaluation} 
\noindent We provide details for the algorithm on splitting and mixing source and labeled target images to evaluate LTS-Mix approach. The algorithm consists of four data mixing configurations and generates two new source images at each iteration. 

\begin{algorithm}[htbp]
	\caption{Mixing source and target patches function}
	\label{algo_mixing}
	\begin{small}
	\textbf{require}: $\{x_s\}$: source image; $\{y_s\}$: source label map; $\{x_s^l\}$: target image; $\{y_t^l\}$: target label map; iter: training iteration;\\
	\texttt{split\_patch}: function that divides the input image into 4 patches of equal size;\\
	\texttt{mix\_patch}: function that groups the 4 input patches into 1 image; \\
	\texttt{concat}: function that concatenates images along the batch axis;   \\
	\textbf{output}: images batch: $x_{st}$; labels batch: $y_{st}$
	\begin{algorithmic}[1]
		\If{iter mod 4 = 0}
		    \State $x_{st} \gets$ $\texttt{concat}(x_s,x_t)$ ;
		     \State $y_{st} \gets$ \texttt{concat}$(y_s,y_t)$ 
		\Else 
		    \State $x_{S1},x_{S2},x_{S3},x_{S4} \gets$ \texttt{split\_patch}$(x_s)$ ;
		    \State $y_{S1},y_{S2},y_{S3},y_{S4} \gets$ \texttt{split\_patch}$(y_s)$ ;
		    \State $x_{T1},x_{T2},x_{T3},x_{T4} \gets$ \texttt{split\_patch}$(x_t)$ ;
		    \State $y_{T1},y_{T2},y_{T3},y_{T4} \gets$ \texttt{split\_patch}$(y_t)$ ;
		    \If{iter mod 4 = 1}
		        \State $x_{st}^1 \gets $ \texttt{mix\_patch}$(x_{S1},x_{T2},x_{S3},x_{T4})$ ;
		        \State $y_{st}^1 \gets $ \texttt{mix\_patch}$(y_{S1},y_{T2},y_{S3},y_{T4})$ ;
		        \State $x_{st}^2 \gets $ \texttt{mix\_patch}$(x_{T1},x_{S2},x_{T3},x_{S4})$ ;
		        \State $y_{st}^2 \gets $ \texttt{mix\_patch}$(y_{T1},y_{S2},y_{T3},y_{S4})$ ;
		    \ElsIf{iter mod 4 = 2}
		        \State $x_{st}^1 \gets $ \texttt{mix\_patch}$(x_{S1},x_{T2},x_{T3},x_{S4})$ ;
		        \State $y_{st}^1 \gets $ \texttt{mix\_patch}$(y_{S1},y_{T2},y_{T3},y_{S4})$ ;
		        \State $x_{st}^2 \gets $ \texttt{mix\_patch}$(x_{T1},x_{S2},x_{S3},x_{T4})$ ;
		        \State $y_{st}^2 \gets $ \texttt{mix\_patch}$(y_{T1},y_{S2},y_{S3},y_{T4})$ ;
		    \ElsIf{iter mod 4 = 3}
		        \State $x_{st}^1 \gets $ \texttt{mix\_patch}$(x_{S1},x_{S2},x_{T3},x_{T4})$ ;
		        \State $y_{st}^1 \gets $ \texttt{mix\_patch}$(y_{S1},y_{S2},y_{T3},y_{T4})$ ;
		        \State $x_{st}^2 \gets $ \texttt{mix\_patch}$(x_{T1},x_{T2},x_{S3},x_{S4})$ ;
		        \State $y_{st}^2 \gets $ \texttt{mix\_patch}$(y_{T1},y_{T2},y_{S3},y_{S4})$ ;
		    \EndIf
		    \State $x_{st} \gets $\texttt{concat}$(x_{st}^1,x_{st}^2)$ ; $y_{st} \gets$\texttt{concat}$(y_{st}^1,y_{st}^2)$ ;
		\EndIf
		\State \textbf{return:} $x_{st}$ and $y_{st}$
	\end{algorithmic}
	\end{small}
\end{algorithm}

\null\newpage

\section{Additional Experiments: Random Seed Sensitivity}
\vspace{-0.5cm}   
\begin{table}[H]
\setlength{\tabcolsep}{4pt}
\setlength\aboverulesep{0pt}\setlength\belowrulesep{0pt}
\caption{Results on GTA5 $ \rightarrow$ Cityscpes for different $N_t^l$ values and random seeds achieved by our proposed methods. Best mIoU per row is in bold and the last best is in italic.}
\vspace{-0.25cm}  
\begin{center}
\small
\begin{tabular}{|c|c|ccccc|cc|}
\cline{3-7}
\multicolumn{2}{c}{} & \multicolumn{5}{|c|}{\texttt{seed}} & \multicolumn{1}{c}{} & \multicolumn{1}{c}{} \\
\hline
$N_t^l$ & Method & 10 & 20 & 30 & 40 & 50 & Mean & Std\\
\hline 
\multirow{5}{*}{100} & ST & 42.52 & 44.11 &	44.03 & \rlast{39.96} &	\rfirst{45.01} &	43.13 &	1.77 \\
\cline{2-9}
& Baseline & \rfirst{51.69} & 50.94 & 50.38 & \rlast{48.26} & 48.71 & 50 & 1.31 \\
& LTT & 48.55 & 49.59 &	\rfirst{51.94} & \rlast{48.18} & 49.14 & 49.48 & 1.32 \\
& LTS & \rfirst{53.04} & 52.82 & 53 & \rlast{51.62} & 52.2 & 52.54 & 0.55 \\
& LTS - Mix & \rfirst{55.2} & 53.9 & 53.76 & \rlast{52.69} & 53.16 & 53.74 & 0.85\\
\hline
\multirow{5}{*}{200} & ST & 48 & 48.79 & 48.3 & \rlast{47.74} & \rfirst{49.63} &	48.49 &	0.67 \\
\cline{2-9} 
& Baseline & 53.1 & \rfirst{53.7} & 51.59 & \rlast{50.77} & 52.98 & 52.43 & 1.08 \\
& LTT & \rlast{51.31} & 51.71 & 51.49 & 51.41 & \rfirst{53.26} & 51.84 & 0.72 \\
& LTS & \rlast{54.12} & 54.41 & 54.32 & 55.38 & \rfirst{56.24} & 54.89 & 0.8 \\
& LTS-Mix & 57.31 & 57.43 & 57.92 & \rlast{57.29} & \rfirst{58.49} & 57.69 & 0.46\\
\hline
\multirow{5}{*}{500} & ST & 55.22 & \rfirst{56.18} & \rlast{54.08} & 54.57 & 54.79 &	54.97 & 0.71 \\
\cline{2-9} 
& Baseline & 55.63 & 56.11 & \rfirst{56.82} & \rlast{55.22} & 55.83 & 55.92 & 0.53 \\
& LTT & 55.42 & 56.09 & \rfirst{56.41} & \rlast{54.68} & 55.55 & 55.63 & 0.6\\
& LTS & 59.11 & 59.01 & \rfirst{59.34} & \rlast{57.77} & 58.5 & 58.75 & 0.56 \\
& LTS-Mix & \rlast{61.4} & \rfirst{62.49} & 61.65 & 61.8 & 62.1 & 61.89 & 0.38\\
\hline
\multirow{5}{*}{1000} & ST & 60.13 & \rfirst{60.19} & \rlast{58.9} & 60.09 & 60.02 & 59.87 & 0.49 \\
\cline{2-9} 
& Baseline & 59.29 & \rfirst{59.61} & 58.8 & \rlast{58.09} & 58.22 & 58.8 & 0.59 \\
& LTT & 59.12 & 59.03 & \rlast{58.57} & \rfirst{60} & 58.82 & 59.11 & 0.48 \\
& LTS & \rfirst{62.03} & \rlast{60.61} & 61.32 & 61.53 & 60.71 & 61.24 & 0.53 \\
& LTS-Mix & 65.14 & 65.36 & 64.58 & \rfirst{65.47} & \rlast{64.24} & 64.96 & 0.47\\
\hline
\end{tabular}
\end{center}
\label{table:results_gta_city_seeds} 
\setlength{\tabcolsep}{1.pt}
\end{table}

\vspace{-0.5cm}  
\begin{table}[H]
\setlength{\tabcolsep}{4pt}
\setlength\aboverulesep{0pt}\setlength\belowrulesep{0pt}
\caption{Results on SYNTHIA $ \rightarrow$ Cityscpes for different $N_t^l$ values and random seeds achieved by our proposed methods. Best mIoU per row is in bold and the last best is in italic.}
\vspace{-0.25cm}
\begin{center}
\small
\begin{tabular}{|c|c|ccccc|cc|}
\cline{3-7}
\multicolumn{2}{c}{} & \multicolumn{5}{|c|}{\texttt{seed}} & \multicolumn{1}{c}{} & \multicolumn{1}{c}{} \\
\hline
$N_t^l$ & Method & 10 & 20 & 30 & 40 & 50 & Mean & Std\\
\hline 
\multirow{5}{*}{100} & ST & 54.69 & \rfirst{56.6} & 56.31 & \rlast{51.04} & 55.12 & 54.75 & 1.99 \\
\cline{2-9}
& Baseline & \rlast{55.2} & 57.5 & \rfirst{60.17} & 56.44 & 56.13 & 57.09 & 1.71 \\
& LTT & \rfirst{57.67} & 55.16 & 56.52 & \rlast{54.62} & 54.63 & 55.72 & 1.2	\\
& LTS & 58.95 & 59.6 & \rfirst{59.86} & \rlast{57.83} & 58.32 & 58.91 & 0.76 \\
& LTS - Mix & 59.46 & 61.61 & \rfirst{61.88} & \rlast{58.48} & 61.08 & 60.5 & 1.31\\
\hline
\multirow{5}{*}{200} & ST & \rfirst{60.94} & 59.33 & 60.21 & 56.04 & \rlast{55.11} &	58.33 & 2.32 \\
\cline{2-9} 
& Baseline & \rlast{57.94} & 59.69 & 60.62 & \rfirst{61.3} & 60.09 & 59.93 & 1.13 \\
& LTT & 59.22 & \rfirst{60.07} & 58.56 & 57.71 & \rlast{57.67} & 58.65 & 0.92 \\
& LTS & \rlast{60.94} & \rfirst{62.39} & 61.85 & 61.9 & 61.54 & 61.72 & 0.48 \\
& LTS-Mix & \rlast{63.7} & 64.52 & \rfirst{64.75} & 64.06 & 64.74 & 64.35 & 0.41\\
\hline
\multirow{5}{*}{500} & ST & \rfirst{66.53} & 65.87 & 65.97 & 64.58 & \rlast{64.29} & 65.45 & 0.86 \\
\cline{2-9} 
& Baseline & \rfirst{65} & 64.6 & 63.69 & 63.68 & \rlast{62.84} & 63.96 & 0.76 \\
& LTT & \rfirst{64.03} & \rlast{59.52} & 62.94 & 61.23 & 61.54 &	61.85 & 1.54 \\
& LTS & 65.28 & \rlast{64.29} & \rfirst{66.68} & 66.08 & 65.67 & 65.6 & 0.8 \\
& LTS-Mix & 69.79 & 70.58 & 70.68 & \rlast{69.74} & \rfirst{70.7} & 70.3 & 0.44\\
\hline
\multirow{5}{*}{1000} & ST & \rfirst{68.76} & 68.53 & \rlast{67.68} & 68.59 & 68.36 & 68.38 & 0.37 \\
\cline{2-9} 
& Baseline & 66.12 & 66.84 & 65.75 & \rlast{65.59} & \rfirst{67.52} & 66.36 & 0.72 \\
& LTT & \rfirst{67.32} & 65.19 & 65.13 & 65.27 & \rlast{64.91} & 65.56 & 0.89 \\
& LTS & 68.21 & \rlast{67.45} & \rfirst{68.95} & 68.41 & 68.87 & 68.38 & 0.54 \\
& LTS-Mix & \rlast{72.09} & \rfirst{73.69} & 72.98 & 72.97 & 72.58 & 72.86 & 0.53\\
\hline
\end{tabular}
\end{center}
\label{table:results_synthia_city_seeds} 
\setlength{\tabcolsep}{1.pt}
\end{table}

\end{document}